\documentclass{article}
\usepackage{amsmath,graphicx,mlspconf}
\usepackage{amssymb}
\usepackage{xcolor}
\usepackage[T1]{fontenc}
\usepackage{hyperref}
\usepackage{multirow}
\usepackage{booktabs}
\usepackage{subcaption}

\copyrightnotice{979-8-3503-2411-2/25/\$31.00 {\copyright}2025 IEEE}

\toappear{2025 IEEE International Workshop on Machine Learning for Signal Processing, Aug.\ 31-- Sep.\ 3, 2025, Istanbul, Turkey}

\title{The Cow of Rembrandt\\Analyzing Artistic Prompt Interpretation in Text-to-Image Models}

\name{Alfio Ferrara, Sergio Picascia, Elisabetta Rocchetti}
\address{Department of Computer Science, Università degli Studi di Milano, Via Celoria, 18, 20133 Milan, Italy}

\begin{document}

\maketitle

\begin{abstract}
Text-to-image diffusion models have demonstrated remarkable capabilities in generating artistic content by learning from billions of images, including popular artworks. However, the fundamental question of how these models internally represent concepts, such as content and style in paintings, remains unexplored. Traditional computer vision assumes content and style are orthogonal, but diffusion models receive no explicit guidance about this distinction during training. In this work, we investigate how transformer-based text-to-image diffusion models encode content and style concepts when generating artworks. We leverage cross-attention heatmaps to attribute pixels in generated images to specific prompt tokens, enabling us to isolate image regions influenced by content-describing versus style-describing tokens. Our findings reveal that diffusion models demonstrate varying degrees of content-style separation depending on the specific artistic prompt and style requested. In many cases, content tokens primarily influence object-related regions while style tokens affect background and texture areas, suggesting an emergent understanding of the content-style distinction. These insights contribute to our understanding of how large-scale generative models internally represent complex artistic concepts without explicit supervision. We share the code and dataset, together with an exploratory tool for visualizing attention maps at \url{https://github.com/umilISLab/artistic-prompt-interpretation}.
\end{abstract}

\section{Introduction}
\label{sec:intro}

\begin{figure}[ht!]
    \centering
    \includegraphics[width=\linewidth]{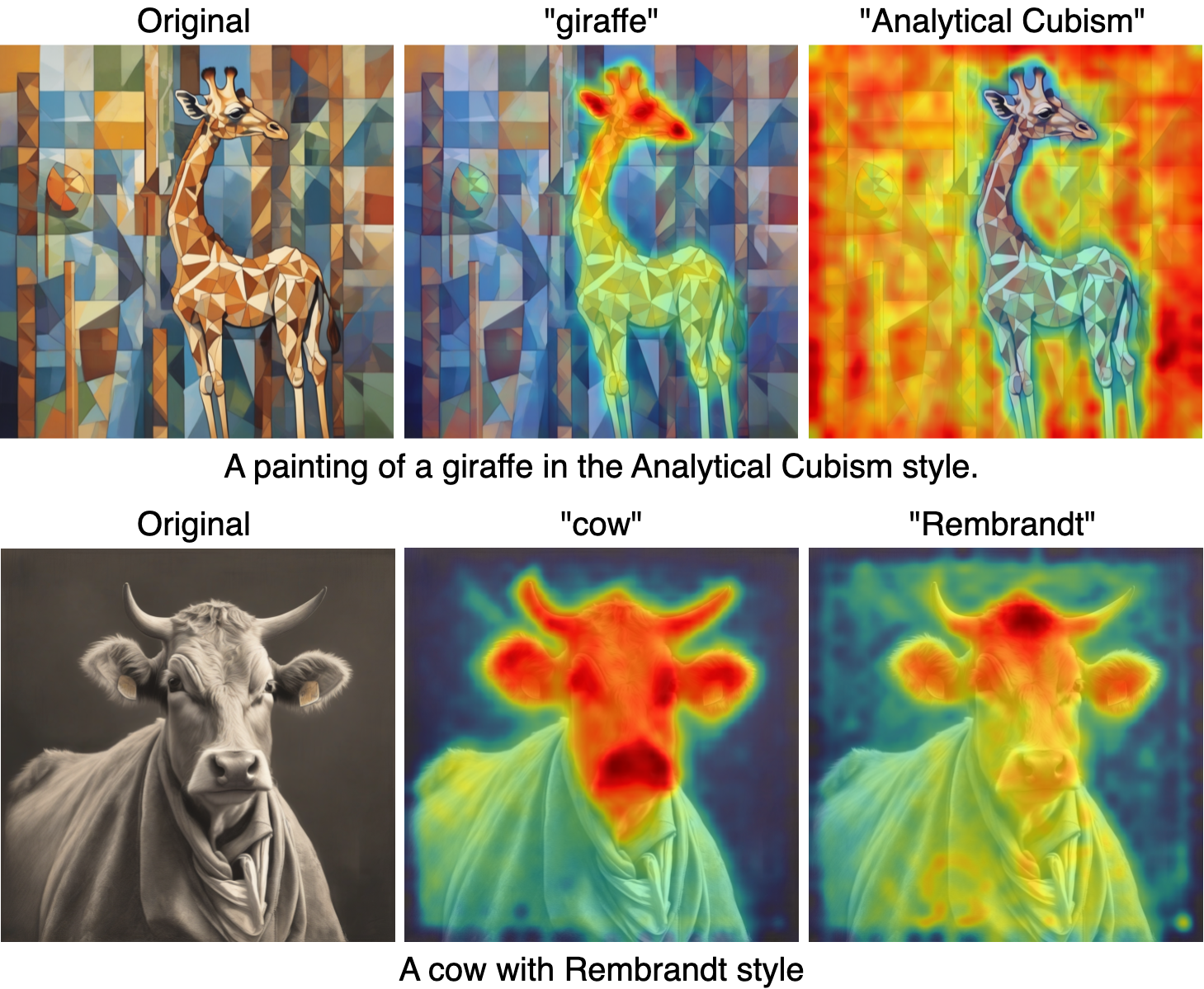}
    \caption{\small (Top left) image generated by a txt2img model using the prompt ``\textit{A painting of a giraffe in the Analytical Cubism style.}'' with the corresponding heatmaps for (top center) content and (top right) style components. (Bottom) image and heatmaps from ``\textit{A cow with Rembrandt style.}''}
    \label{fig:example-giraffe}
\end{figure}

Nowadays, we employ models that take a textual prompt as input and generate an image as output, an image which, in most cases, closely reflects the description provided in the prompt. These text-to-image (\textit{txt2img}) models are trained on billions of images sourced from the internet, including artworks from open-access, labeled online repositories such as WikiArt~\cite{salehLargescaleClassificationFineArt2015}. Through this training, the models implicitly acquire knowledge of art, artists, and artistic movements~\cite{somainiAlgorithmicImagesArtificial2023}. This learned knowledge is often leveraged when the model is instructed, for instance, to generate ``\textit{a painting of a giraffe in the Analytical Cubism style.}'' This leads to intriguing questions: How does the \textit{txt2img} model interpret stylistic instructions? And how does it represent the interplay between the content to depict and the style to apply?

Historically, the field of Computer Vision (CV) has treated the notions of content and style as orthogonal, assuming that content is independent of style, and vice versa~\cite{wuNotOnlyGenerative2023}. However, such a distinction is not explicitly provided to diffusion models during training. Moreover, disentangling content from style goes beyond simply associating content with \textit{what} is depicted and style with \textit{how} it is depicted. This motivates a deeper investigation into how these models internally conceptualize and separate these elements.

In this work, we focus on analyzing the behavior of transformer-based txt2img diffusion models when generating artworks, with a particular emphasis on how they encode the concepts of \textit{style} and \textit{content}. To achieve this, we leverage the models’ cross-attention heatmaps, which allow us to attribute each pixel in the generated image to specific tokens in the input prompt. This allows to isolate image regions primarily influenced by tokens describing content, and those influenced by tokens describing style. If these two regions overlap significantly, it may suggest that the model interprets content and style as similar or entangled; if they are distinct, it may indicate that the model conceptually separates the two. Figure~\ref{fig:example-giraffe} shows the cross-attention heatmaps corresponding to the words ``\textit{giraffe}'' and ``\textit{Analytical Cubism}'', which define the content and style in the prompt ``\textit{A painting of a giraffe in the Analytical Cubism style}''. In this example, the two heatmaps highlight largely complementary regions: the content component ``\textit{giraffe}'' activates areas corresponding to the animal itself, while the style component ``\textit{Analytical Cubism}'' influences the rest of the image. This suggests a separation of influence between content and style components during image generation. However, this distinction is not evident in the bottom example of Figure~\ref{fig:example-giraffe}, which reveals an interesting behavior: the model appears to apply Rembrandt’s style by ``dressing'' the cow, since part of the activated region associated with the style component includes the cow’s clothing.

The remainder of this paper is organized as follows: Section~\ref{sec:relwork} reviews foundational concepts and related work in computer vision and content–style disentanglement. Section~\ref{sec:method} outlines our methodology for investigating the model’s understanding of content and style. Section~\ref{sec:setup} describes our experimental setup and configurations. Section~\ref{sec:results} presents our results and offers interpretations of how the model distinguishes (or conflates) content and style. Finally, Section~\ref{sec:conclusions} concludes the paper and discusses potential directions for future research.

\section{Related Work}
\label{sec:relwork}
In this section, we begin by introducing key concepts related to txt2img models and their underlying mechanisms. We then review relevant literature on the interpretability of these models. Finally, we introduce the task of content–style disentanglement and discuss prior work that addresses this challenge.

\textbf{txt2img generation and diffusion models.} Our work centers on the study of txt2img diffusion models. These models learn to predict the amount of noise that has been progressively added to an image, transforming it into Gaussian noise during training. At inference time, the model reverses this process to generate an image from noise only~\cite{NEURIPS2020_4c5bcfec}. In txt2img generation, diffusion models are conditioned on textual prompts using text encoders. A standard architecture for such models typically includes a language encoder (e.g., CLIP~\cite{radfordLearningTransferableVisual2021}) for generating word embeddings, a variational autoencoder (VAE)~\cite{kingma2022autoencodingvariationalbayes} to translate between latent representations and images, and a time-conditional U-Net~\cite{UNET} for denoising latent vectors~\cite{diffusionrombach2022}.

\textbf{Interpretability of txt2img models.} Given the increasing use and impact of these models, there is a growing need for interpretability tools tailored to txt2img diffusion architectures. Among the various approaches offered by the field of Explainable AI, our focus lies on cross-attention-based methods. These models rely on cross-attention mechanisms to connect the textual information embedded in input prompts with the visual content generated during denoising. This mechanism enables the generation of heatmaps that attribute regions of the image to specific words in the prompt. Several recent works leverage this property to explain diffusion-based models by producing cross-attention attribution maps~\cite{tangWhatDAAMInterpreting2023, helblingConceptAttentionDiffusionTransformers2025, stanLVLMInterpretInterpretabilityTool2024}. These maps can be used to investigate how textual prompts shape visual outputs, analyze syntactic structures through head–dependent heatmap interactions, and explore semantic phenomena such as feature entanglement. Additionally, attention-based explanations have proven valuable for downstream tasks such as object segmentation, as they visually delineate what the model perceives as coherent objects~\cite{tangWhatDAAMInterpreting2023,  Tian_2024, Wang_2025, ma2023diffusionseg}. Beyond segmentation, attention maps also facilitate the discovery of biases and inconsistencies in model behavior~\cite{stanLVLMInterpretInterpretabilityTool2024}.

\textbf{Content-style disentanglement.} Distinguishing between style and content in art remains a longstanding challenge in the field of Computer Vision (CV)~\cite{wuNotOnlyGenerative2023}. While content—the subject or objects depicted in an artwork—is typically straightforward to identify, style is far more elusive. There is no universally accepted definition of style, and its components—such as color usage, brushstrokes, composition, and perspective~\cite{graham-2012, somepalliMeasuringStyleSimilarity2024}—often overlap with content, making disentanglement difficult~\cite{wuNotOnlyGenerative2023}. Moreover, the notion of style can vary depending on the specific CV task. 
Previous work has tackled this challenge by attempting to disentangle the internal representations of models, aiming to separate the latent space into distinct content and style components~\cite{wuNotOnlyGenerative2023, kotovenko2019content, kazemi2019style, gabbay2020improving}. 

While these approaches primarily focus on generating images that match a target style, our goal is different. In this work, we aim to investigate how txt2img diffusion models perceive the artistic concepts of content and style by analyzing attention heatmaps. This approach allows us to probe the internal mechanisms of these models, offering insights into how they distinguish—or conflate—style and content.

\section{Methodology}
\label{sec:method}

The objective of this study is to quantify how transformer-based txt2img diffusion models distinguish between content and style concepts when generating paintings. Specifically, we analyze the spatial distribution of cross-attention values assigned to image pixels from content and style tokens in the conditioning prompt. We systematically construct prompts with varying content and style components, then analyze the spatial distribution of cross-attention values between these tokens and image pixels during generation. To measure content-style separation, we threshold attention maps into binary masks and compute their Intersection over Union (IoU), comparing this overlap against baseline token overlap to determine whether the model spatially distinguishes content from style or treats these concepts as entangled. Due to the need for internal representations, our work only applies to open-source txt2img models.

\textbf{Prompt Templates.} When generating images with txt2img models, prompts typically contain various modifiers that influence the output~\cite{oppenlaenderTaxonomyPromptModifiers2024a}. To maintain experimental clarity and focus on our research question, we specifically isolated content and style components as variables in our prompt templates, deliberately excluding additional modifiers that might confound analysis. This approach aligns with established methodologies in the prompt engineering literature~\cite{liuDesignGuidelinesPrompt2022}, which inspired us in the selection of four distinct prompt templates, to ensure robust results regardless of the specific phrasing: (i) \textit{a painting of a \texttt{<CONTENT>} in the \texttt{<STYLE>} style}; (ii) \textit{a \texttt{<STYLE>} painting of a \texttt{<CONTENT>}}; (iii) \textit{a \texttt{<CONTENT>} in the \texttt{<STYLE>} style}; (iv) \textit{a \texttt{<CONTENT>} with \texttt{<STYLE>} style}. 

Taking these four templates, \texttt{<CONTENT>} should be replaced with any subject to be depicted in the generated image (e.g., a physical object, person, landscape, etc.), and \texttt{<STYLE>} should be replaced with the name of an artist or artistic movement that defines the desired stylistic approach (e.g., Picasso, Art Nouveau, etc.).

\textbf{Heatmap Extraction.} To quantify the influence of specific prompt tokens on generated image regions, we employ Diffusion Attentive Attribution Maps (DAAM) following the methodology established by Tang et al.~\cite{tangWhatDAAMInterpreting2023}. This technique enables us to extract token-specific attribution maps that visualize the spatial relationship between textual concepts and visual elements. Formally, we analyze the denoising network of the diffusion model, typically implemented as a U-Net architecture comprising $i$ sequential downsampling and upsampling convolutional blocks. These blocks compute an internal representation $h_{i,t}$ at every time step $t$. The internal representations are then conditioned on the word embeddings $X$ of the prompt using multi-headed cross-attention layers. 

$F^{(i)}_t[x, y,l, k]$ is the cross-attention array, normalized to $[0, 1]$, connecting the $k$-th word to the intermediate coordinates $(x, y)$ for the $i$-th downsampling block and $l$-th head. To generate a unified attribution map, we aggregate cross-attention arrays across spatial, temporal, and architectural dimensions. First, we normalize all intermediate attention arrays through bicubic interpolation to match the original image dimensions $(w,h)$. We then compute the token-specific attribution map $D^R_k$ for the $k$-th token as:

\begin{equation}
    D^R_k[x,y] = \sum_{t_j, i,l}(\tilde{F}^{(i)}_{t_j,k,l}[x,y])
\end{equation}

The resulting $D^R_k$ is a normalized heatmap with values in $[0, 1]$, where higher intensities indicate stronger attribution between the $k$-th token and the corresponding image pixel at coordinates $(x, y)$.




\textbf{IoU Evaluation.} To quantitatively analyze the spatial relationships between content and style representations, we transform the continuous attribution heatmaps into binary segmentation masks by applying a threshold parameter $\tau$ to each normalized attribution map $D^R_k$: 
\begin{equation}
    D^{\mathbb{I}_\tau}_k[x,y] = \mathbb{I}(D^R_k[x,y] \geq \tau)
\end{equation}
where $\mathbb{I}(\cdot)$ denotes the indicator function, yielding a binary mask where pixels with attention values exceeding $\tau$ are set to $1$, and all others to $0$. For each generated image, we compute the Intersection over Union (IoU) between the content token mask $D^{\mathbb{I}_\tau}_C$ and the style token mask $D^{\mathbb{I}_\tau}_S$:

\begin{equation}
    \text{IoU}_{\text{CS}}=\frac{|D^{\mathbb{I}_\tau}_C \cap D^{\mathbb{I}_\tau}_S|}{|D^{\mathbb{I}_\tau}_C \cup D^{\mathbb{I}_\tau}_S|}
\end{equation}

To establish whether content and style tokens exhibit distinctive spatial attention patterns or tend to focus on overlapping image regions, we compare $\text{IoU}_{\text{CS}}$ against a baseline metric, $\text{mIoU}_{\text{B}}$. The baseline defines the mean IoU computed across all possible token pairs in the prompt, with the constraint that each pair includes exactly one token from the set $\{C, S\}$ (either content or style) and one token from the remaining tokens in the prompt. We use this comprehensive baseline since we cannot exclude a priori lexical categories, e.g. stopwords, because they can still carry valuable information~\cite{bai2025identifyinganalyzingperformancecriticaltokens}; therefore, including all tokens provides an unbiased measure of the general attention landscape. This baseline serves to characterize the general overlap in attention across the prompt: a high $\text{mIoU}_{\text{B}}$ suggests that tokens generally attend to similar image regions, whereas a low value indicates more spatially differentiated attention patterns. Given the variability in $\text{mIoU}_{\text{B}}$ across different prompts, we introduce a metric to quantify the deviation of $\text{IoU}_{\text{CS}}$ from the baseline. This difference metric $\Delta = \text{mIoU}_{\text{B}} - \text{IoU}_{\text{CS}}$ serves as our primary indicator: a positive $\Delta$ value suggests that content and style tokens attend to distinct spatial regions (lower mutual overlap compared to other pairs), while a negative $\Delta$ indicates that content and style tokens exhibit higher spatial correlation than expected, suggesting conceptual entanglement in the internal representations of the model. We employ a paired \textit{t}-test to determine statistical significance in the difference between $\text{IoU}_{\text{CS}}$ and $\text{mIoU}_{\text{B}}$. 

\section{Experimental setup}
\label{sec:setup}
In this section, we describe how we construct the prompts to generate images, the chosen txt2img model and the choices of threshold $\tau$ for binary heatmap construction.

\textbf{Prompt construction.} To populate the templates described in Section \ref{sec:method} with diverse and representative content elements, we utilize the 80 object class labels from the MS COCO dataset~\cite{linMicrosoftCOCOCommon2014}. For style components, we incorporate a comprehensive collection of 50 style descriptors from the WikiArt dataset~\cite{salehLargescaleClassificationFineArt2015}, comprising 23 individual artists and 27 artistic movements. The combinatorial expansion of these templates, content classes, and style descriptors yields a substantial experimental corpus of 16,000 unique prompts, which are subsequently processed by a txt2img model.

\textbf{Image Generation.} We generate images using the Stable Diffusion XL\footnote{\url{https://huggingface.co/stabilityai/stable-diffusion-xl-base-1.0}}~\cite{sdxl} (SDXL) text2img model, with 30 inference steps. SDXL is one of the most established open-source txt2img models available on the HuggingFace platform.

\textbf{Threshold $\tau$.} We explored two strategies for selecting the most prominent regions in the heatmaps. The first strategy employs a fixed absolute threshold $\tau$, applied directly to the continuous values of $D_k^R$. We evaluated a range of thresholds, specifically $\tau \in \{0.1, 0.2, \dots, 0.9\}$. This approach aligns more closely with the method used in~\cite{tangWhatDAAMInterpreting2023}, where a fixed threshold of $\tau = 0.4$ was adopted for downstream analysis. The second strategy adopts a relative threshold, where $\tau$ is set to the $p$-th percentile value of the $D_k^R$ distribution, with $p \in \{0.1, 0.2, \dots, 0.9\}$. This percentile-based method adapts to the characteristics of the heatmaps generated by a specific txt2img model, and guarantees a minimum percentage support of $1 - p$ for the sets used in the IoU computation.

\section{Results}
\label{sec:results}

In this section, we present both quantitative and qualitative results illustrating how SDXL behaves when selecting between content and style tokens as sources of information during the image generation process.

\begin{figure}[ht!]
    \centering
    \includegraphics[width=0.8\linewidth]{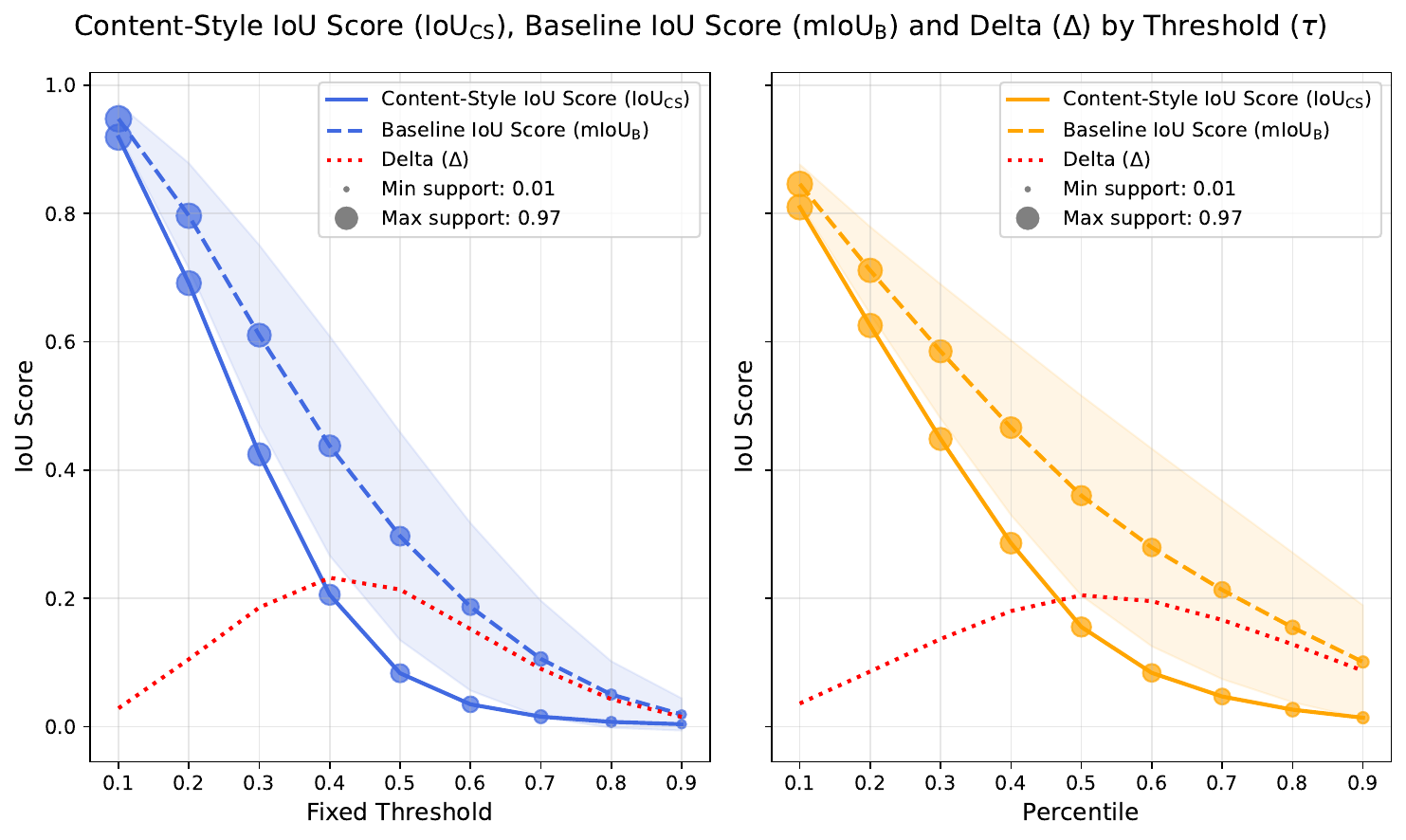}
    \caption{\small Content-Style IoU score ($\text{IoU}_{\text{CS}}$, solid line) is compared against the Baseline IoU score ($\text{mIoU}_{\text{B}}$, dashed line) and their difference ($\Delta$, dotted line) across varying threshold values ($\tau$). Markers on the IoU curves are sized proportionally to the mean support of the corresponding sets used in the computation. In the left plot, thresholds are fixed; in the right plot, thresholds are derived from percentile values based on the $D_k^R$ distribution.}
    \label{fig:scores-by-threshold}
\end{figure}

\textbf{Quantitative evaluation.} Figure~\ref{fig:scores-by-threshold} illustrates how $\text{IoU}_{\text{CS}}$, $\text{mIoU}_{\text{B}}$, and their difference $\Delta$ vary across different threshold ($\tau$) configurations. Notably, $\text{IoU}_{\text{CS}}$ remains consistently and significantly lower than $\text{mIoU}_{\text{B}}$, particularly for fixed thresholds and percentile-based thresholds between $0.2$ and $0.6$. This suggests that, on average, the heatmap regions corresponding to content and style tokens do not substantially overlap. This observation is further supported by statistical analysis: we perform paired \textit{t}-tests and measure the standardized distance between their means in terms of standard deviations. For all threshold configurations, the null hypothesis of equal means can be rejected with $p$-values below $0.001$, while, on average, $\text{IoU}_{\text{CS}}$ lies $0.64$ standard deviations away from $\text{mIoU}_{\text{B}}$. From now on, we comment results obtained setting a fixed $\tau = 0.4$.

Our analysis reveals that the distribution of $\Delta$ remains stable across different prompt templates but exhibits substantial variation based on the specific content and style tokens employed. Table~\ref{tab:mean-deltas} presents the highest and lowest average $\Delta$ values observed for both content and style components. We recall that high $\Delta$ values indicate scenarios where content and style components direct attention to spatially disjoint regions within the generated image.

Examination of content components with elevated $\Delta$ values reveals a predominance of animal-related terms, whereas \textit{person} exhibits the lowest value. Within style components representing artists, we find the only occurrence of a negative mean $\Delta$: the Dutch painter Rembrandt. Furthermore, our analysis demonstrates that figurative art movements correlate with higher $\Delta$ values, while abstract art movements correspond to lower $\Delta$ values, suggesting a systematic relationship between artistic representation style and attention distribution patterns.

\begin{table}[ht!]
\centering
\resizebox{\linewidth}{!}{%
\begin{tabular}{@{}llllll@{}}
\toprule
\multicolumn{2}{c|}{\textbf{Content}}                  & \multicolumn{2}{c|}{\textbf{Style (Artist)}}                    & \multicolumn{2}{c}{\textbf{Style (Movement)}} \\ \midrule
giraffe    & \multicolumn{1}{l|}{0.43 \textit{(0.08)}} & Durer     & \multicolumn{1}{l|}{0.35 \textit{(0.10)}}  & New Realism            & 0.39 \textit{(0.08)} \\
banana     & \multicolumn{1}{l|}{0.39 \textit{(0.09)}} & Sargent    & \multicolumn{1}{l|}{0.33 \textit{(0.10)}}  & Rococo                 & 0.38 \textit{(0.09)} \\
sheep      & \multicolumn{1}{l|}{0.37 \textit{(0.12)}} & Dali      & \multicolumn{1}{l|}{0.29 \textit{(0.09)}}  & Minimalism             & 0.35 \textit{(0.09)} \\
bear       & \multicolumn{1}{l|}{0.37 \textit{(0.11)}} & Konchalovsky & \multicolumn{1}{l|}{0.29 \textit{(0.11)}}  & Ukiyo e                & 0.35 \textit{(0.08)} \\
zebra      & \multicolumn{1}{l|}{0.35 \textit{(0.12)}} & Monet       & \multicolumn{1}{l|}{0.27 \textit{(0.11)}}  & Art Nouveau            & 0.34 \textit{(0.09)} \\
...      & \multicolumn{1}{l|}{} & ...       & \multicolumn{1}{l|}{}  & ...            &  \\
cell phone & \multicolumn{1}{l|}{0.10 \textit{(0.17)}} & Kustodiev    & \multicolumn{1}{l|}{0.15 \textit{(0.11)}}  & Impressionism          & 0.21 \textit{(0.17)} \\
tie        & \multicolumn{1}{l|}{0.06 \textit{(0.17)}} & Pissarro   & \multicolumn{1}{l|}{0.14 \textit{(0.15)}}  & Cubism                 & 0.13 \textit{(0.18)} \\
hot dog    & \multicolumn{1}{l|}{0.04 \textit{(0.08)}} & van Gogh   & \multicolumn{1}{l|}{0.12 \textit{(0.17)}}  & Abstract Expr. & 0.11 \textit{(0.18)} \\
bed        & \multicolumn{1}{l|}{0.04 \textit{(0.15)}} & Kirchner   & \multicolumn{1}{l|}{0.09 \textit{(0.13)}}  & Expressionism          & 0.09 \textit{(0.16)} \\
person     & \multicolumn{1}{l|}{0.03 \textit{(0.09)}} & Rembrandt          & \multicolumn{1}{l|}{-0.07 \textit{(0.17)}} & Baroque                & 0.07 \textit{(0.14)} \\ \bottomrule
\end{tabular}%
}
\caption{\small Highest and lowest average $\Delta$ values for type of components. Standard deviation reported in parenthesis.}
\label{tab:mean-deltas}
\end{table}

\begin{figure*}[ht!]
    \centering
    \begin{subfigure}[t]{0.3\textwidth}
        \centering
        \includegraphics[width=\textwidth]{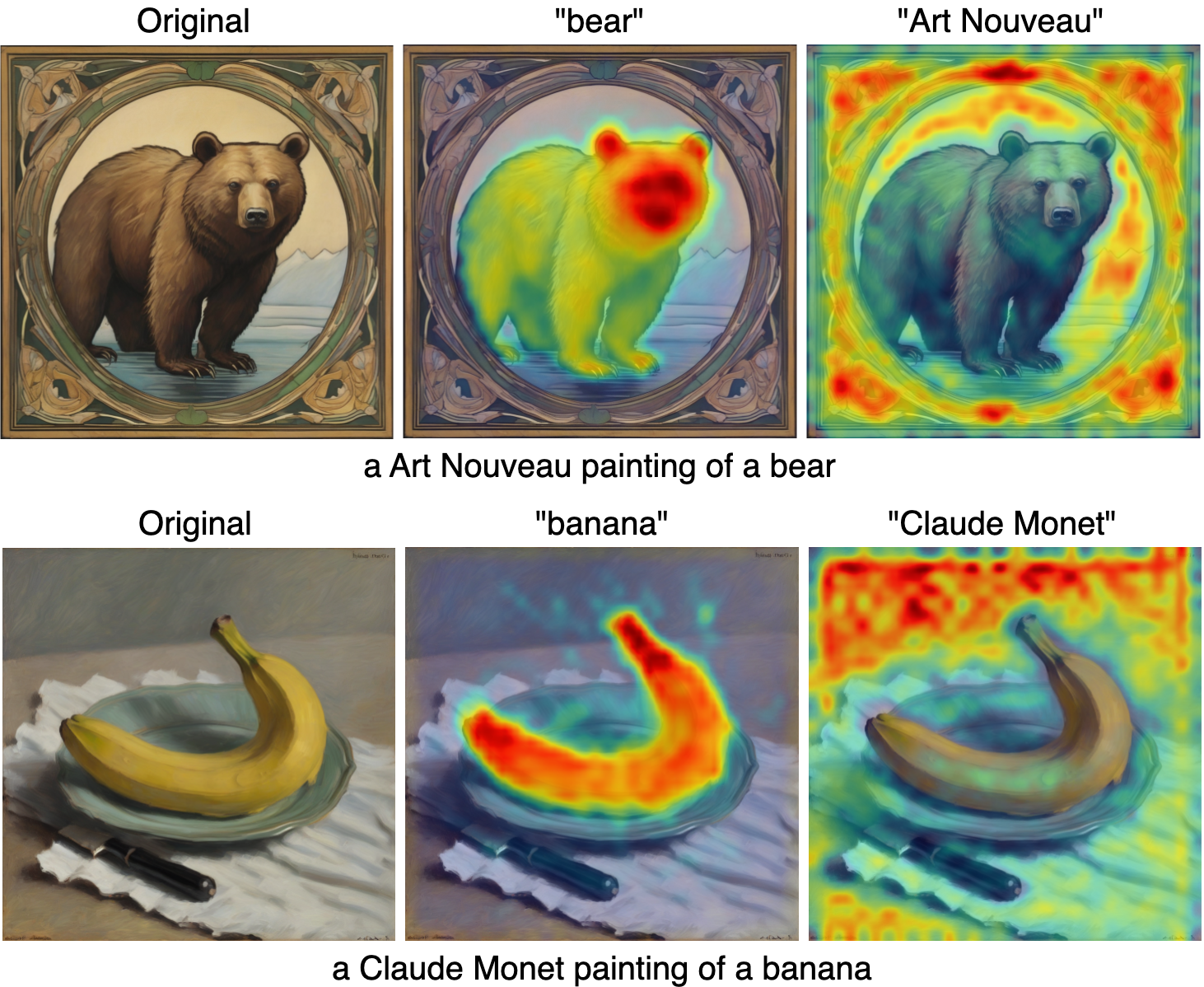}
        \caption{}
        \label{fig:examples-top}
    \end{subfigure}%
    ~ 
    \begin{subfigure}[t]{0.3\textwidth}
        \centering
        \includegraphics[width=\textwidth]{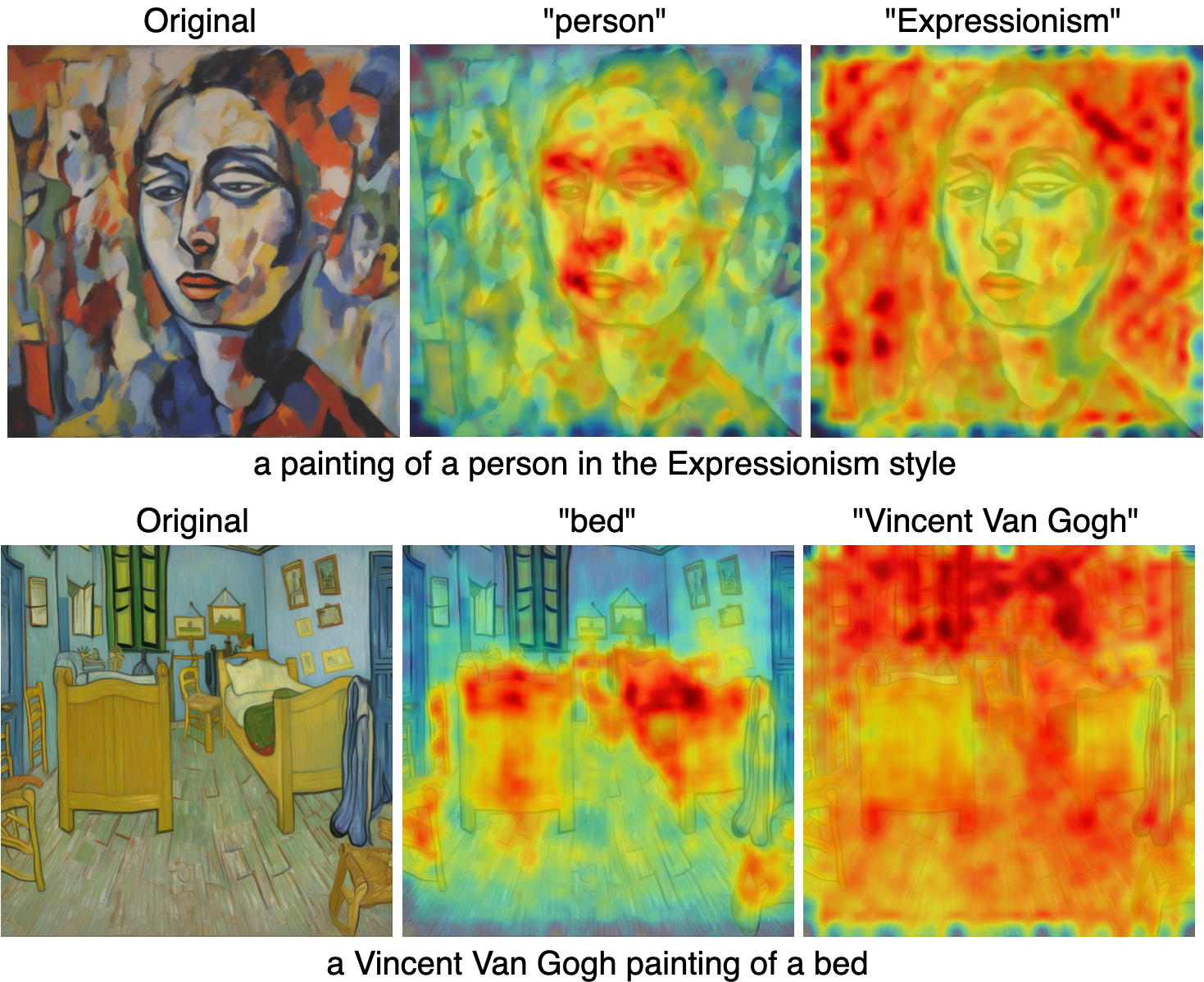}
        \caption{}
        \label{fig:examples-bottom}
    \end{subfigure}
    ~ 
    \begin{subfigure}[t]{0.3\textwidth}
        \centering
        \includegraphics[width=\textwidth]{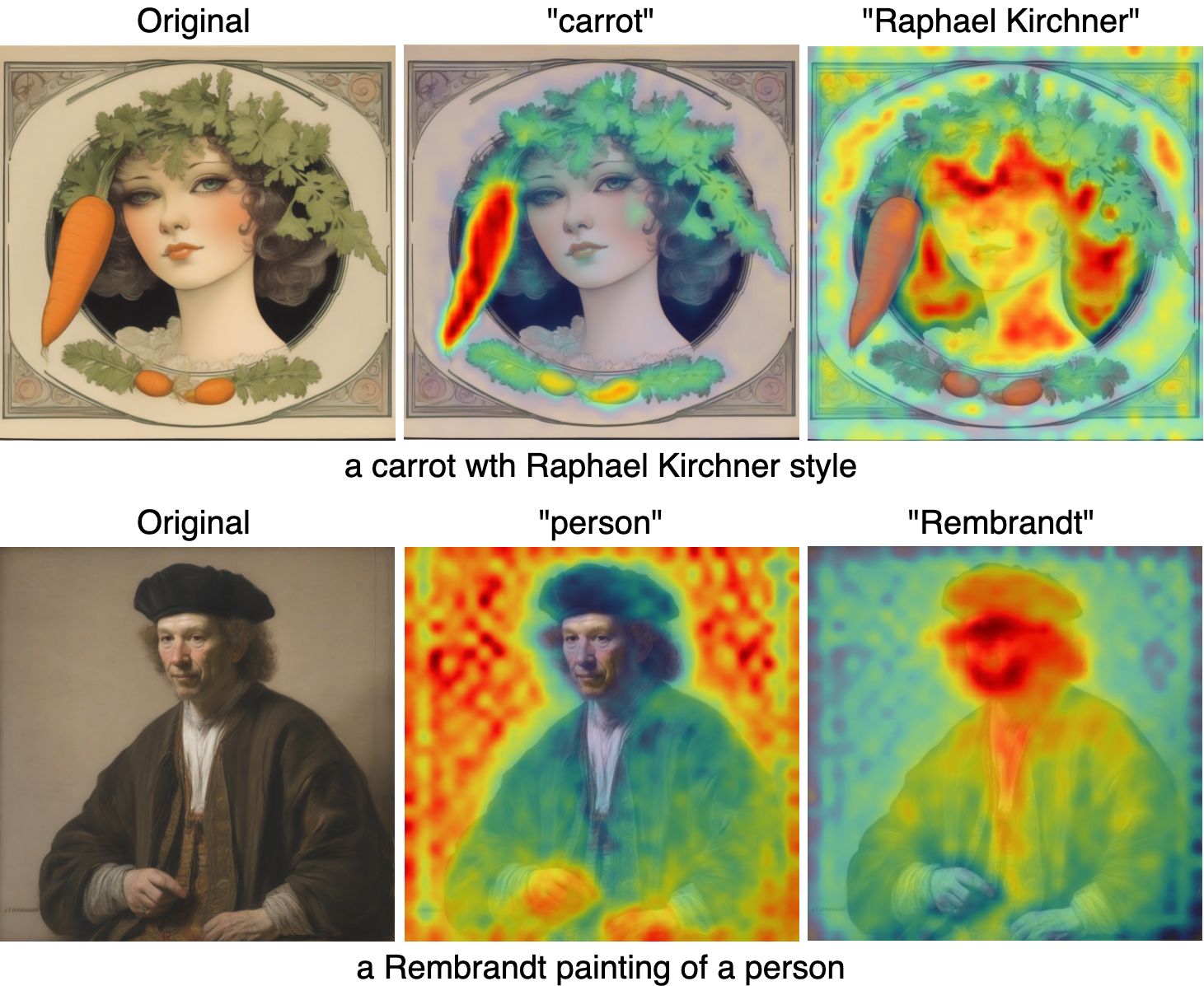}
        \caption{}
        \label{fig:examples-particular}
    \end{subfigure}
    \caption{\small Examples of generated images with clearly distinct content and style components (a), with overlapping content and style components (b), with edge cases (c).}
    \label{fig:examples}
\end{figure*}

\textbf{Qualitative evaluation.} Figure~\ref{fig:examples} illustrates representative examples of generated images with their corresponding content and style attention heatmaps.  On the left, Figure~\ref{fig:examples-top} shows examples of clear spatial separation between content and style components, representing the predominant behavior observed throughout our study. Conversely, at the center, Figure~\ref{fig:examples-bottom} exhibits cases with low $\Delta$ scores, indicating substantial overlap between content and style attention regions. Notably, in these overlapping cases, the primary subject matter appears integrated with the background rather than distinctly separated as in the previous example. Figure~\ref{fig:examples-particular}, on the right, reveals intriguing edge cases in model behavior. In the upper example, the \textit{carrot} generated with \textit{Raphael Kirchner}'s style is accompanied by a female figure, consistent with Kirchner's predominant subject matter of women in his paintings. The lower example presents a rare instance of positive $\Delta$ values observed in a painting generated with \textit{Rembrandt} style. This behavior is observed when the \textit{Rembrandt} token is used in combination with the content \textit{person}, contrasting with the typically low $\Delta$ values observed for this component in all the other cases. This anomaly likely stems from Rembrandt's extensive self-portraiture, causing the \textit{Rembrandt} style token to attend strongly to facial features and human forms.


\section{Conclusions}
\label{sec:conclusions}
This work investigates how transformer-based txt2img diffusion models convey the concepts of content and style when generating paintings. We analyse IoU scores computed on DAAM corresponding to the content and style components in the input prompt. Results indicate that, on average, content and style components tend to influence complementary regions of the generated images. The degree of separation is non-trivial and varies primarily depending on the specific content and style terms used. We identify and analyze edge cases—instances where $\Delta$ values are particularly high or low—and provide illustrative examples. Furthermore, we highlight cases suggesting that the model’s training data can significantly affect this behavior: subjects that appear frequently in stylistic contexts may be internalized by the model as stylistic elements themselves.

Future work will extend this analysis by exploring alternative methodological choices, such as variations in cross-attention extraction, heatmap resolution, and overlap metrics, and by evaluating carefully selected content and style components that more closely reflect realistic artistic scenarios. We plan to expand our investigation to other txt2img models and are pursuing collaborations with domain experts for human evaluation and deeper analysis of results.


\bibliographystyle{IEEEbib}
\bibliography{refs}

\end{document}